\documentclass{article} % For LaTeX2e
\usepackage{iclr2026_conference,times}

\usepackage{amsmath,amsfonts,bm}

\def\eqref#1{equation~\ref{#1}}
\def\1{\bm{1}}

\DeclareMathAlphabet{\mathsfit}{\encodingdefault}{\sfdefault}{m}{sl}
\SetMathAlphabet{\mathsfit}{bold}{\encodingdefault}{\sfdefault}{bx}{n}

\DeclareMathOperator*{\argmax}{arg\,max}

\usepackage{hyperref}
\usepackage{url}
\usepackage{algorithm}
\usepackage{algpseudocode}
\usepackage{graphicx}
\usepackage{subcaption}
\usepackage{makecell} % add this in your preamble
\usepackage{wrapfig} % put this in the preamble

\usepackage[utf8]{inputenc}   % allows UTF-8 characters in the source
\usepackage[T1]{fontenc}      % better font encoding for accented and special characters
\usepackage{textcomp}         % extra symbols

\title{On the relationship between the choice of \\ representation and in-context learning}

\author{Ioana Marinescu \\ NYU \\\texttt{im2178@nyu.edu} \And Kyunghyun Cho \\NYU \& Genentech \\\texttt{kc119@nyu.edu} \And  Eric Karl Oermann \\ NYU\\\texttt{oermae01@nyu.edu}}

\iclrfinalcopy % Uncomment for camera-ready version, but NOT for submission.
\begin{document}

\maketitle

\begin{abstract}
In-context learning (ICL) is the ability of a large language model (LLM) to learn a new task from a few demonstrations presented as part of the context. Past studies have attributed a large portion of the success of ICL to the way these in-context demonstrations are represented, particularly to how labels are represented in classification tasks. On the other hand, observations of the learning capacity of ICL (i.e., the extent to which more in-context demonstrations can lead to higher performance) have been mixed, and ICL is often thought to occur only under specific conditions. The interaction between these two aspects in ICL, representation and learning, has not been studied in depth until now. We hypothesize that they are largely independent of one another, such that the representation of demonstrations determines the baseline accuracy of ICL, while learning from additional demonstrations improves only on top of this baseline. We validate this hypothesis by developing an optimization algorithm that can enumerate a spectrum of possible label sets (representations) varying in semantic relevance. We then perform ICL with varying numbers of in-context demonstrations for each of these label sets. We observed that learning happens regardless of the quality of the label set itself, although its efficiency, measured by the slope of improvement over in-context demonstrations, is conditioned on both the label set quality and the parameter count of the underlying language model. Despite the emergence of learning, the relative quality (accuracy) of the choice of a label set (representation) is largely maintained throughout learning, confirming our hypothesis and implying their orthogonality. Our work reveals a previously underexplored aspect of ICL: the independent effects of learning from demonstrations and their representations on ICL performance.

\end{abstract}

\section{Introduction}
LLMs are able to learn a new task from a few examples, an ability known as in-context learning (ICL) \citep{NEURIPS2020_1457c0d6, dong-etal-2024-survey}. A model is prompted with input-output pairs (demonstrations) illustrating the task and then asked to make a prediction for a novel input. 
The ICL paradigm is appealing as the models appear to learn something new without updating any weights, in contrast with the typical way in which a neural network learns via backpropagation. However, the performance of ICL depends heavily on properties of the given demonstrations \citep{NEURIPS2021_5c049256}, such as the the distribution of input text, the label space \citep{min-etal-2022-rethinking}, the number and order of examples \citep{Lu2021FantasticallyOP, liu2024letslearnstepstep, chen-etal-2023-many, bertsch-etal-2025-context}, and the overall format of the sequence \citep{Zhao2021CalibrateBU}. It remains unclear whether ICL truly constitutes learning, and if so, how learning interacts with elements of the prompt. 

Prior work has studied learning and representation in ICL separately, not considering the interaction between the two, which may have led to incomplete conclusions. 
According to earlier studies, different kinds of in-context learning happen depending on the choice of how labels are represented. In particular, two types of labeling schemes have been studied extensively: gold (or semantically-meaningful) labeling and abstract (or semantically-void) labeling.  
\cite{Pan2023WhatIL} found that with an abstract set of labels, smaller models perform similarly regardless of how many demonstrations were presented, while larger models showed increased performance with more demonstrations. This led them to conclude that the emergence of in-context learning depends on the model size. More recently, \cite{kirsanov-etal-2025-geometry} observed that LLMs are sensitive to the representation of labels and perform better with gold labels than with abstract labels. In their study, the accuracy improved with an increasing number of demonstrations for both gold and abstract labels, even with a smaller model. 
Both \cite{min-etal-2022-rethinking} and \cite{Pan2023WhatIL} observed that breaking the input-output correspondence while preserving the set of labels had a minimal effect for small models, suggesting that the representation is the sole driver of performance, rather than the demonstration pairings themselves. These findings highlight the need to investigate the interaction between learning and representation in ICL.

\begin{figure}
    \centering
    \includegraphics[width=\linewidth]{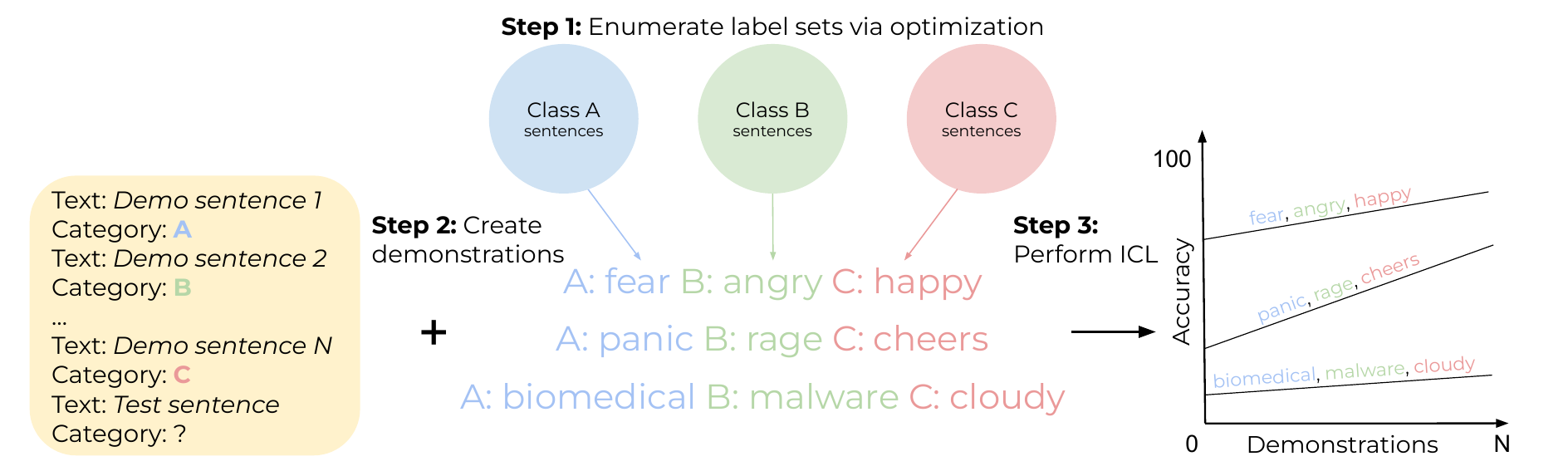}
    \caption{Method overview. Step 1: We develop an optimization algorithm to enumerate a list of possible label sets for a sentiment classification task. Step 2: We label demonstration sentences according to the label sets found. Step 3: We use these demonstrations in ICL tasks and evaluate the performance obtained with each label set on the same set of test sentences. }
    \label{fig:fig1}
\end{figure}

In this work, we propose that in classification tasks ICL performance is influenced by two separate components: representation - the choice of class names or labels, and learning - the number of examples presented in context. To quantify the role of representation, we evaluate label sets with varying degrees of semantic relevance to the task. We develop an optimization algorithm to enumerate such label sets. We then use these representations to label input sentences and to create demonstrations for ICL. We conduct experiments on a sentiment classification task: 3-way and 5-way, across three model sizes. 
For each label set we analyze ICL performance while varying the number of demonstrations. We show an overview of our method in Figure~\ref{fig:fig1}. 

We found that representation steers learning, although learning typically happens regardless of representation and model size. The {\it ranking} of representations in terms of accuracy is constant across different number of demonstrations, following the initial order (without any demonstrations). Moreover, the accuracy range attainable with a given representation is largely determined by the zero-shot accuracy.
For most label sets, the $N$-shot accuracy generally increases with $N$, although we found that learning efficiency, that is the slope of improvement, depends on the model size. 
This characterization of the relationship between learning and representation in ICL suggests that it is possible to improve ICL performance by carefully choosing an appropriate label set representation for the task. Our code is available at \url{https://github.com/ioanam25/class-representation-icl}. 
% allows us to design better prompts that maximize ICL performance
% choose class names that maximize learning. 

% Our optimization algorithm can find semantically-meaningful label sets for any classification task, eliminating the need for annotating datasets with gold labels and for prompt optimization techniques that are highly dependent on the prompt rather than the overall task. 

\section{Related work}

There have been a flurry of academic studies on ICL that have revealed its properties and characterized ICL as a new class of learning, since \citet{NEURIPS2020_1457c0d6} demonstrated the (surprising) effectiveness of ICL with a large-scale language model. In this section, we list up some of these studies that have shed light on ICL over the past few years.

\paragraph{Content effects.} 

Recent studies suggest that LLMs are not fully-abstract reasoners, that is, they do not always learn a function which they can apply to an arbitrary input \citep{10.1093/pnasnexus/pgae233}. Instead, these models show content effects similar to those of humans who reason more accurately about familiar or grounded situations, compared to unfamiliar or abstract ones. \cite{doi:10.1073/pnas.2322420121} found that LLM accuracy is influenced by the probability of the task to be performed, the probability of the target output, and the probability of the provided input. The bias towards outputs that have a high prior probability occurs in ICL as well. LLMs do not always identify a unique input-output mapping across the demonstrations, in order to apply it to the test input. They rely instead on the combination of their prior knowledge and presented demonstrations. There are several factors influencing ICL, such as the order \citep{Lu2021FantasticallyOP} and number of demonstrations \citep{chen-etal-2023-many}, input and output distributions, and the overall format of the prompt~\citep{min-etal-2022-rethinking}. 
According to these studies, ICL may ignore the task defined by the demonstrations and instead resort to using the prior obtained from pretraining. This implies that ICL may not be considered learning under a strict definition, wherein learning must capture the input-output correspondence in a given training set.

% \cite{Pan2023WhatIL} disentangle ICL into task recognition (TR), which recognizes the task from demonstrations and applies LLMs’ pre-trained priors, and task learning (TL), which learns a new input-label mapping from demonstrations. They find that gold labels and random labels (uniformly sample a label from the class labels to assign it to an input) in the demonstration result in similar accuracy. Using random labels does not preserve the input-output mapping, so the model does not even have a chance to learn it, thus this claim about prior probability of the correct label does not reveal any information about learning since the label space (set of possible classes) is the same. Another finding here suggests that TL emerges with larger models and more demonstrations, but for not smaller models. In particular, they find that using an abstract set of labels (which preserve the input-output mapping) underperforms random labels. We believe this is insufficient to show that TL is not present in smaller models, as these are different label spaces, and again, the random labeling does not obey the mapping, so it's not representative of what a model can learn from the context (in this case, the mapping). In this work, we propose that TL is, in fact always happening, and is independent of the representation of the task, and it scales with the number of demonstrations.\\

\paragraph{Learning mechanisms.} 

Theoretical work has explained ICL as implicit Bayesian inference by training language models from scratch on controlled synthetic data \citep{xie2022an, NEURIPS2023_73950f0e, panwar2024incontext, jiang2023latentspacetheoryemergent}. \cite{arora2025bayesian} have shown that Bayesian scaling laws are a good fit for the ICL curve. Another line of studies has interpreted ICL as implicitly performing gradient descent~\citep{pmlr-v202-von-oswald23a, NEURIPS2023_8ed3d610} and/or other types of learning algorithms \citep{akyurek2023what, NEURIPS2022_c529dba0, NEURIPS2023_b2e63e36, pmlr-v202-li23l}. All these mathematical observations encourage us to view ICL as a real learning algorithm and to perform careful empirical investigations to study its properties in real-world settings.

\paragraph{Pretraining data distribution.} 

ICL is known to emerge from pretraining when the pretraining data, or its distribution, exhibits a particular set of properties.
\cite{chan2022datadistributionalpropertiesdrive} found that ICL emerges when data exhibits burstiness (items appear in clusters rather than being uniformly distributed over time) and follows a skewed Zipfian distribution. \cite{NEURIPS2023_2e10b2c2} identified a task diversity threshold during pretraining beyond which language models can perform well on unseen ICL regression tasks. \cite{hahn2023theoryemergentincontextlearning} found that ICL arises from generic next-token prediction when the pretraining distribution has a sufficient amounts of compositional structure. 

% make some conclusion on first part: so all these say that ICL actually learns something when models are big and many demonstrations but not in other cases. what we're saying is that the model is ALWAYS learning something about the input output mapping and it's not a matter of number of demonstrations (we see it in both few shot and many shot - there's no threshold)/model size, but rather of the representation of the label space and is influenced by the zero-shot accuracy of the correct label for the test input.\\ 

\paragraph{Prompt optimization.} 
By deepening our theoretical understanding of the interaction between representation and learning, we can further improve ICL. A common approach to improving LLMs' performance without any extra weight update is via ``prompt engineering,'' that is, by crafting prompts manually. Recent studies introduce prompt optimizers that search over strings to identify high-performing prompts \citep{yuksekgonul2025optimizing, zhou2023large, yang2024large, guo2024connecting, agrawal2025gepareflectivepromptevolution}. These approaches typically optimize one prompt at a time. For ICL classification tasks, we propose a method to optimize the class names on a separate ``labeling'' set of sentences, and directly use them as labels in new ICL prompts.

\section{Method}

\subsection{In-context learning formulation}

We formulate the goal of an ICL task as solving
\begin{align}
    \argmax\limits_{y \in \mathcal{C}}p(\tau(y)|x, D_{\tau}),
\end{align}
where $D_{\tau}=\left\{(x_n, \tau(y_n)) \right\}_{n=1}^N$ refers to a (small) number of input-output pairs. $\tau(y)$ defines a label set or how we represent each class $y \in \left\{ 1, 2, \ldots, \mathcal{C} \right\}$ as a token in a predefined vocabulary, i.e., $\tau: \left\{ 1, 2, \ldots, \mathcal{C} \right\} \to V$, where $V$ is a vocabulary of unique tokens. $D_{\tau}$ refers to presenting the dataset $D$ using $\tau$ to encode the classes. By properly formatting $D$, $x$ and $\tau(y)$, LLMs have been found to be able to implicitly learn to predict the correct label associated with a new instance $x$.

Prior work has observed that ICL achieves better performance with gold labels than with abstract labels \citep{Pan2023WhatIL}. For example, \cite{kirsanov-etal-2025-geometry} analyzed a sentiment classification task. The model performed better on an ICL task with gold labels such as \{\textit{joy, anger, fear}\} than with abstract labels such as \{\textit{A, B, C}\}, even if the input-output correspondence was the same for both label sets.

While abstract labels lead to worse performance than gold labels, the accuracy increases with more examples for either of the label sets. Based on this observation, and taking into account the content effects revealed by \cite{10.1093/pnasnexus/pgae233}, we 
% propose that in-context learning is sensitive to label semantics. 
% ICL works as a combination of two separate components:
propose to factor ICL's predictive probability into the product of two probabilities:
\begin{align}
    p(\tau({y})|x, D_{\tau}) \propto q(\tau(y)|x, D_{\tau}) p(\tau({y})|x).
\end{align}
The first term $q(\tau{(y)}|x,D)$ corresponds to {\it learning}, and the second term $p(\tau{(y)}|x)$ corresponds to {\it prior} knowledge learned by the language model during pretraining. We assume that the first component, {\it learning}, is largely invariant to how we represent the classes. In other words,
\begin{align}
    q(\tau(y) | x, D_{\tau}) \approx q(\tau'(y) | x, D_{\tau'}).
\end{align}

On the other hand, the prior knowledge must be sensitive to the choice of $\tau$, as it lacks the context which is presented in the form of in-context demonstrations. Unless $\tau(y)$ is {\it meaningful} under the pretraining corpus, the language model cannot work with an arbitrary representation of a class {\it a priori}. That is, it is almost certain that
\begin{align}
    p(\tau(y) | x) \neq p(\tau'(y) | x),
\end{align}
for $\tau \neq \tau'$. 

In this work, we investigate how the contributions of learning and and prior knowledge are disentangled in ICL. 
% Given this theoretical proposal of how ICL works, we can choose $\tau$ properly in order to maximally benefit from the in-context learning capability of LLMs. 
We design a readily actionable way to find a good label map $\tau$ systematically, in order to facilitate this investigation.

\subsection{Class representation optimization}

We describe a systematic method to choose a label set $\tau$ that will maximize the performance of ICL across any set of inputs from the same task family. For example, for a sentiment classification task, we can find optimal labels for the classes, and then use these labels as the outputs in ICL demonstrations (input-output pairs) for any other set of inputs.

We assume access to a set of $K$ examples, which we refer to as a labeling set, and knowledge of the class that each example belongs to (how the examples are clustered). The goal is to find, for each class, a name, that is represented by a single token in the vocabulary, that is meaningful under the pretraining corpus. To name $\mathcal{C}$ classes, we want to choose a set of ${\mathcal{C}}$ tokens from $|V|$ possible tokens in a given vocabulary, $\tau = (l_1, l_2,...l_{\mathcal{C}}) \in V^{\mathcal{C}}$. A good representation map $\tau$ should maximize the probability assigned to the correct class $y^\star$, when represented as $\tau(y^\star)$. 
% We expect that the optimal class names will be related to each other and to the topic of the task. 
We can write this directly as an objective function:
\begin{align}
    \max_{(l_1, l_2,...l_{\mathcal{C}}) \in V^{\mathcal{C}}} \sum_{k=1}^K \left(f_{\theta}(x_k, l_{y_k}) - \log \sum_{c=1}^{\mathcal{C}} \exp (f_{\theta}(x_k, l_{c}))\right),
    \label{eq:optimization}
\end{align}
where $x_k$ are the input examples, $y_k \in \{1, 2,...{\mathcal{C}}\}$ are the classes they belong to, $l_{y_k} = \tau(y_k)$ is the label assigned to class $y_k$, and $f_{\theta}$ is the language model's logit. Since the tokens in the label set represent class names and appear after the phrase \textit{``Category:''}, we restrict the vocabulary to tokens that start with the character Ġ (which marks a space and the beginning of a new word).

We optimize this objective via hill climbing, shown in Algorithm~\ref{alg:hill_climb}: we start with an initial random token assignment for each class and iterate the following until no improvements can be made: (1) for each class, try all possible alternative tokens while keeping the rest of class names fixed, (2) evaluate the objective under the current assignment, (3) pick the best token if it improves the overall objective, (4) if there is an improvement, repeat. We run this algorithm ten times while varying random seeds and pick the assignment out of up to ten that maximizes the objective in Equation~\ref{eq:optimization}. 

As $K$, the number of examples used to find a label assignment, increases it becomes harder to find an assignment for which the labels have high probability for many input sentences. To maximize the objective, that assignment should be generalizable: class names should be meaningful for other possible inputs. Thus, as $K$ increases, we expect the semantics of the labels to be closer to those of gold labels. Equivalently, those labels' zero-shot accuracy for new inputs would be higher with larger $K$. By exploiting the dependence of quality on 
$K$, we obtain a diverse set of label groups that vary in their semantic relevance to the given classification task.

\begin{algorithm}
\caption{Hill Climbing for Token Assignment Optimization}
\small
\label{alg:hill_climb}
\begin{algorithmic}[1]
\Require Initial token assignment for each class
\Require Set of candidate tokens, training sentences with labels
\Ensure Optimized token assignments

\Function{HillClimb}{$initial\_assignments$}
    \State $assignments \gets initial\_assignments$
    \State $objective \gets$ \Call{CalculateObjective}{$assignments$}
    
    \Repeat
        \State $improved \gets \texttt{False}$
        
        \For{each $class$ in classes}
            \State $candidates \gets$ all tokens except current token for $class$
            
            \For{each $token$ in $candidates$}
                \State Compute total objective value assigning current token to this class, Eq. \ref{eq:optimization}
            \EndFor
            
            \State $best\_token \gets$ token with highest objective
            
            \If{$best\_token$ improves current objective}
                \State $assignments[class] \gets best\_token$
                \State Update $objective$
                \State $improved \gets \texttt{True}$
                \State \textbf{break} \Comment{Try next class}
            \EndIf
        \EndFor
    \Until{not $improved$ or max iterations reached}
    
    \State \Return $assignments, objective$
\EndFunction
\end{algorithmic}
\end{algorithm}

\section{Experimental setup}

We conduct a series of experiments to test the hypothesis that learning and representations are largely disentangled in ICL. First, we want to test whether learning emerges regardless of the choice of label representation. For this to be true, for any label set, the $N$-shot accuracy should be increasing with $N$. Second, we want to see how representations influence the learning trajectory. For this, we look at how the $N$-shot accuracy relates to the zero-shot accuracy (for the test input) across the different label representations. We conduct experiments with three different size open-weight models: Llama 3.2 1B, Llama 3.1 8B, Llama 3.1 70B Instruct~\citep{grattafiori2024llama3herdmodels}. We first apply the optimization Algorithm~\ref{alg:hill_climb} to obtain a series of label sets with varying quality for a classification task. Then, we sample demonstrations and name the outputs according to the label set. We prompt a model with the relabeled and concatenated demonstrations to evaluate the ICL performance on these new inputs.

\paragraph{Data and prompting.}

We use a synthetic sentiment classification dataset from \cite{kirsanov-etal-2025-geometry}, which contains 1,000 sentences split equally among 5 classes for 5-way classification. We also use a subset of 600 sentences covering only 3 of the classes for 3-way classification. We split the dataset into a labeling set (25\%), a demonstration set (25\%), and a test set (50\%). The labeling set is used to enumerate class name assignments, the demonstration set is used for the support examples for ICL, and the test set is used for the query inputs in ICL. For each $N$-shot classification task, the task is presented in a minimal format with no explicit instructions, only $N$ demonstrations and a query sentence.

\paragraph{Label sets.}

We evaluate different label sets in ICL. These label sets do not break the original input-output correspondence and only replace the original label names, i.e. the assignment of the classes remains the same. Each label set is obtained by optimizing Equation~\ref{eq:optimization} using $K \in \{10, 20,...100\}$ examples. We show the label sets found with each of the three models in Appendix~\ref{appendix:label_sets} Table \ref{tab:labels_3classes} for 3-way classification and Table \ref{tab:labels_5classes} for 5-way classification. The examples used for finding a label set are the same for each fixed $K$ across all model sizes. Some of the $K$ values (adjacent ones) resulted in the same label set.

We illustrate a few of the label sets obtained for 3-way classification with the 70B model. Naturally, using a small $K=10$ leads to overfitting on labels that have a high zero-shot probability only for those labeling examples. This yielded random words as labels such as \{\textit{biomedical, malware, cloudy}\}. With a small $K$, we cannot find label sets that appear relevant for a sentiment classification task. For a medium value of $K=40$, the labels obtained are more general \{\textit{panic, rage, Cheers}\}, a much better fit for the task. While these labels are clearly descriptive, they are slightly odd choices for class names. Finally, using a large $K=70$ leads to natural category names for a sentiment classification task such as \{\textit{fear, angry, happy}\}. 

The label sets obtained with the same $K$ value vary with different models. For instance, for $K=100$, the 1B model found \{\textit{spectacle, dance, condolences, peril, pissed}\}, the 8B model found \{\textit{surprising, joyful, sorrow, fears, anger}\}, and the 70B model found \{\textit{surprise, happy, sad, anxious, ang}\}. In general, the label sets found by larger models appear to be more semantically meaningful.

\paragraph{In-context learning.}

We sample $N \in \{0,1,...40\}$ examples from the demonstration set and name them according to one of the label sets previously obtained. For the 70B model, we only ran experiments with $N \in \{0, 10, 20, 30, 40\}$ due to compute limitations. For the 1B and 8B models, we ran experiments with $N$ up to 100, as shown in Appendix~\ref{appendix:100demos}. We create demonstrations with a given label set by using that set to label the inputs in a context, and preserve the original input-output mapping from the dataset. These input-output pairs are concatenated, and, together with a query, are given as a prompt to a model. The model then predicts the class for a novel input selected from the test set, which has not been shown in any of the demonstrations and was not used to compute the label sets. We report the average accuracy for the test set, over 10 runs, in which the inputs of the demonstrations are resampled every time.  
\section{Results}
\begin{figure}[t]
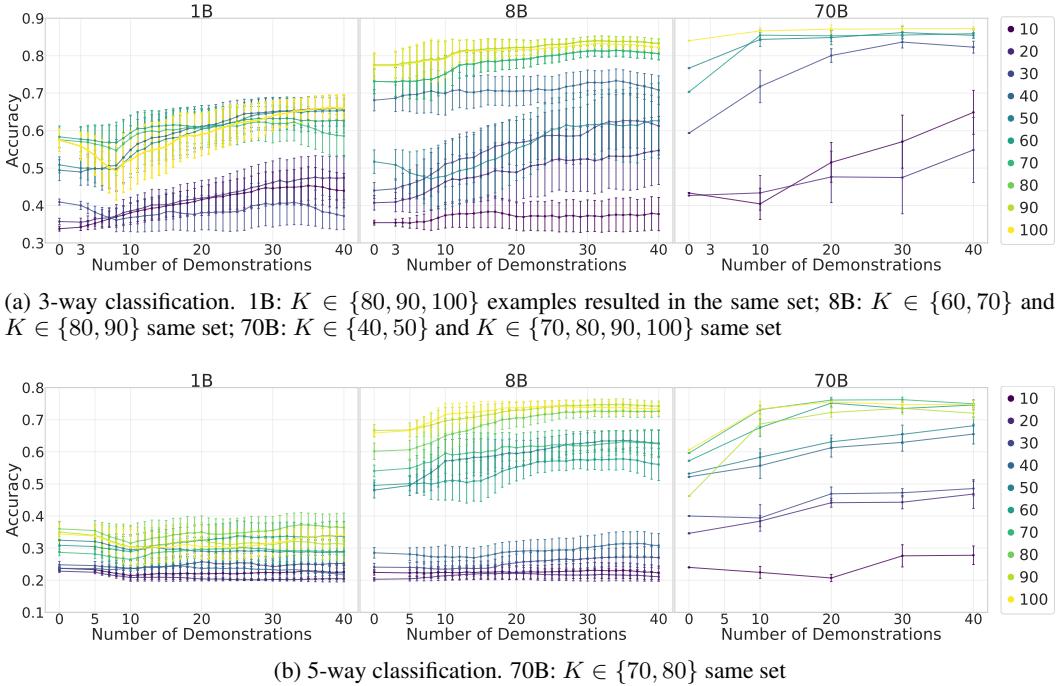

    \centering
    \begin{subfigure}{\linewidth}
        \centering
        \includegraphics[width=\linewidth]{figures/accuracy_3classes.pdf}
        \caption{3-way classification. 1B: $K \in \{80, 90, 100\}$ examples resulted in the same set; 8B: $K\in\{60, 70\}$ and $K\in\{80, 90\}$ same set; 70B: $K\in\{40, 50\}$ and $K\in \{70, 80, 90, 100\}$ same set}
        \label{fig:accuracy_3classes}
    \end{subfigure}
    
    \vspace{1em} % space between the two subfigures

    \begin{subfigure}{\linewidth}
        \centering
        \includegraphics[width=\linewidth]{figures/accuracy_5classes.pdf}
        \caption{5-way classification. 70B: $K\in\{70, 80\}$ same set}
        \label{fig:accuracy_5classes}
    \end{subfigure}

    \caption{Accuracy vs. number of demonstrations across model sizes for (a) 3-class and (b) 5-class settings. The curves were smoothed with a window size of 10, with error bars showing 95\% CI over 10 runs. The legend shows the number of labeling examples $K$ used to fit the label set. Different $K$ values may result in the same label sets. For these sets, the color shown is that of the higher $K$.}
    \label{fig:main_result}
\end{figure}

Figure~\ref{fig:main_result} shows the accuracy vs. number of demonstrations in ICL tasks with different label sets for the 1B, 8B, and 70B models, for 3-way (Figure~\ref{fig:accuracy_3classes}) and 5-way classification (Figure~\ref{fig:accuracy_5classes}). Across all experimental conditions, we observe that the accuracy is generally increasing with the number of demonstrations. There are exceptions, such as when the label set found has a very small zero-shot test accuracy, most curves stay flat, especially for the harder task of 5-way classification. The zero-shot accuracies span a a wide range from chance to ceiling:  33\% to 87\% for 3-way classification and 20\% to 76\% for 5-way classification. The representations with a lower zero-shot accuracy typically resulted from optimization on a small $K$ labeling examples, while those with a high zero-shot accuracy resulted from a larger $K$. The ordering of the label sets as determined by their zero-shot accuracy generally stays constant across $N$-shot tasks, suggesting a consistent ranking of label sets in terms of ICL performance, regardless of the number of demonstrations.

\subsection{Role of representation in ICL}
\begin{figure}[t]
    \centering
    \begin{subfigure}{0.4\textwidth}
        \centering
        \includegraphics[width=\linewidth]{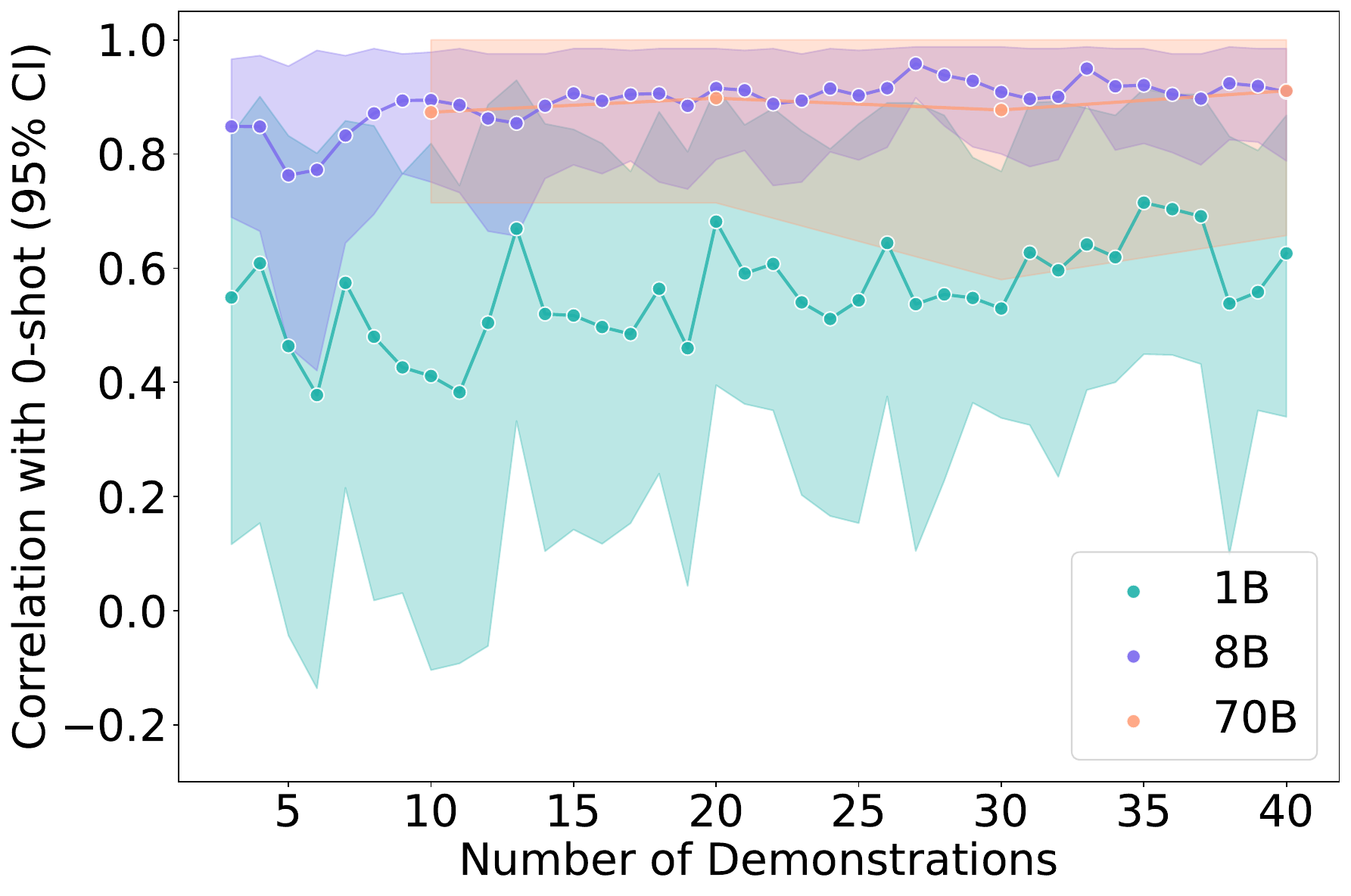}
        \caption{3-way classification}
        \label{fig:vertical_correlations_3classes}
    \end{subfigure}
    \begin{subfigure}{0.4\textwidth}
        \centering
        \includegraphics[width=\linewidth]{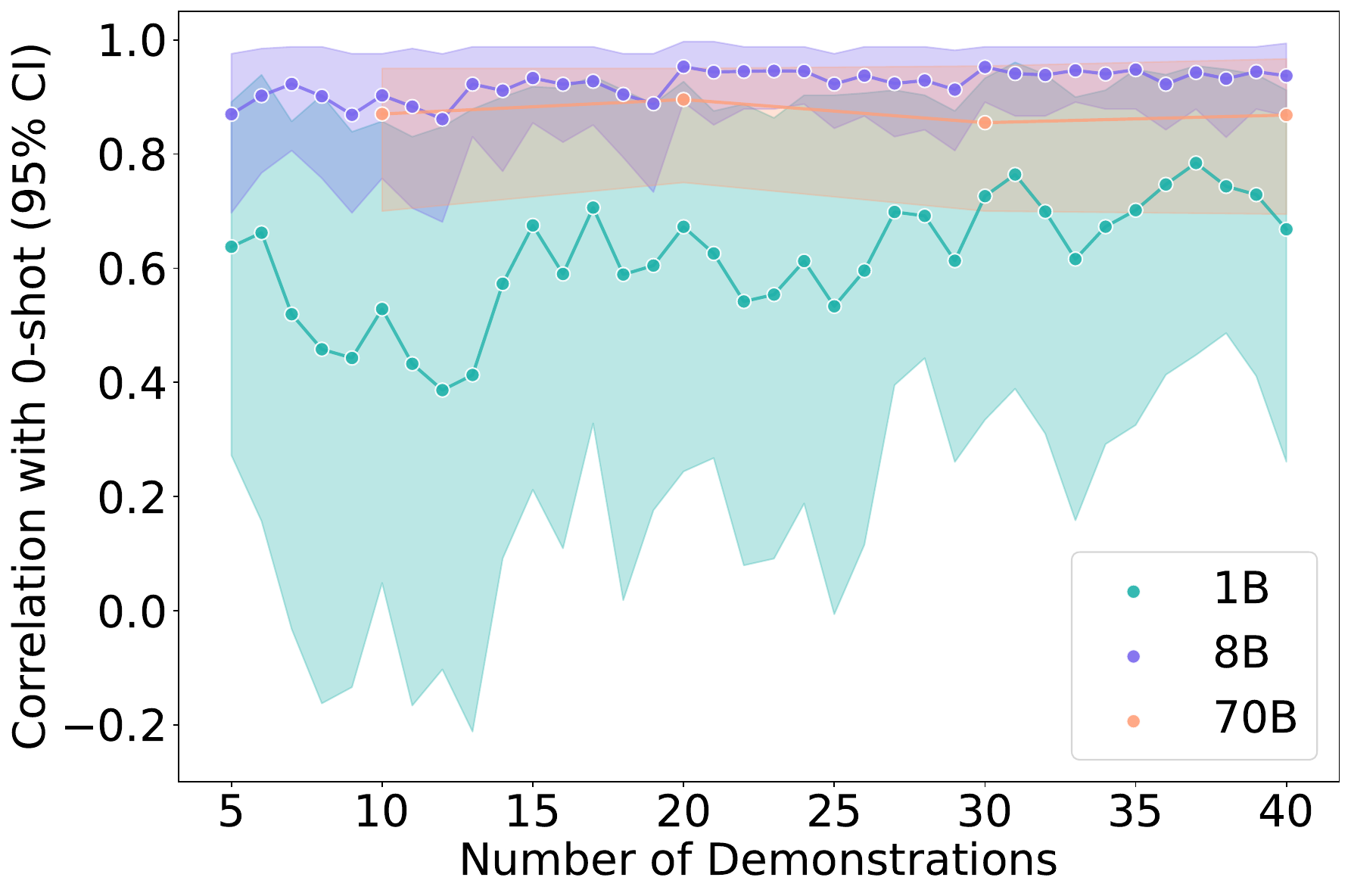}
        \caption{5-way classification}
        \label{fig:vertical_correlations_5classes}
    \end{subfigure}

    \caption{Ranking correlation coefficient between the zero-shot accuracy and the $N$-shot accuracy vs. $N$ number of demonstrations. $N \in \{\text{num classes},...40\}$ for 1B and 8B models, $N \in \{10, 20, 30, 40\}$ for 70B model. The CI are computed over 1000 bootstrapping samples from 10 runs per N-shot accuracy. \textbf{The order of label sets in terms of quality stays consistent across $N$-shot experiments.}}
    \label{fig:vertical_correlations}
\end{figure}

\paragraph{Consistent label set ranking.}

The $N$-shot accuracy of an ICL task using a label set depends on the zero-shot accuracy with that label set: the $N$-shot accuracy is typically higher for label sets with higher zero-shot accuracy and can only grow up to a limit. This is consistent across label sets.
ICL performs better if the label set is meaningful under the pretraining corpus. We observe that for each $N$-shot classification task, the accuracies for ICL with different label sets are ordered according to their initial zero-shot accuracy. We compute the ranking correlation between the zero-shot accuracies and the $N$-shot accuracies (of all the label sets) with $N \in \{\text{num classes},...40\}$ for 1B and 8B models, $N \in \{10, 20, 30, 40\}$ for 70B model. We find that the correlations are indeed high across all model sizes, for both 3-way and 5-way classification (see Figure~\ref{fig:vertical_correlations}), although there is a lot of variance for the 1B model.

% The lowest correlation is obtained in the 5-way classification, for the 1B model. The highest correlation is obtained in the 3-way classification, for the 8B model.

\paragraph{Representation limits the accuracy range.} 

If the zero-shot accuracy of a given label set is low, it is very difficult for ICL to reach a high accuracy regardless of how many demonstrations are used. Reaching a high accuracy with a low zero-shot accuracy label set might require a very large number of demonstrations. Most of the curves appear to increase more slowly around 40 demonstrations, indicating a possible upper bound. The chosen label set thus largely determines the range of accuracies attainable with that representation. However, there are exceptions where the accuracy has not yet plateaued with 40 demonstrations (see Figure~\ref{fig:accuracy_3classes} 70B model, $K=10$), suggesting that it is possible to overcome the limits of the representation with a large number of demonstrations and a larger model. Our findings indicate that the choice of representation is an essential factor when studying ICL and the role of demonstrations, and they shed light on some earlier findings. For example, \cite{Pan2023WhatIL} found that an abstract label set underperformed random allocation of the gold labels to the inputs of the demonstrations, and claimed that this meant that the models could not truly learn the task, but rather relied on their priors. We instead attribute their finding to the fact that the abstract label set has a much lower zero-shot accuracy than a gold label set, and the accuracy increase from learning from additional demonstrations was insufficient to overcome the baseline limitation, which is typically the case for smaller models.

\subsection{When does ICL learn?}

\paragraph{Learning almost always happens.}

We observe that if the zero-shot accuracy is above some threshold, the curves are always increasing regardless of the model size. For the 3-way classification task (Figure~\ref{fig:accuracy_3classes}), the threshold zero-shot accuracy is very low (33\%, chance level), and all curves increase monotonically. For the 5-way classification (Figure~\ref{fig:accuracy_5classes}), the threshold is higher (40\%, double the chance accuracy), and the behavior is more complicated. We analyze it here. For the 1B model, all label sets have a zero-shot accuracy below the threshold, and the learning curves appear flat. It is possible that the increase is small and these models could reach a higher accuracy with a larger number of demonstrations. For the 8B model, some label sets are below the threshold and correspond to flat curves, while some are above and correspond to increasing curves. For the 70B model, the same trends hold with one exception: the label set with a zero-shot accuracy of 35\% ($K=20$), which is on the lower side, has large gains from seeing demonstrations. It appears that with a sufficiently good representation, all models, regardless of size, are able to benefit (to different extents) from more demonstrations.

\begin{figure}[t]
    \centering
    \begin{subfigure}{0.4\textwidth}
        \centering
        \includegraphics[width=\linewidth]{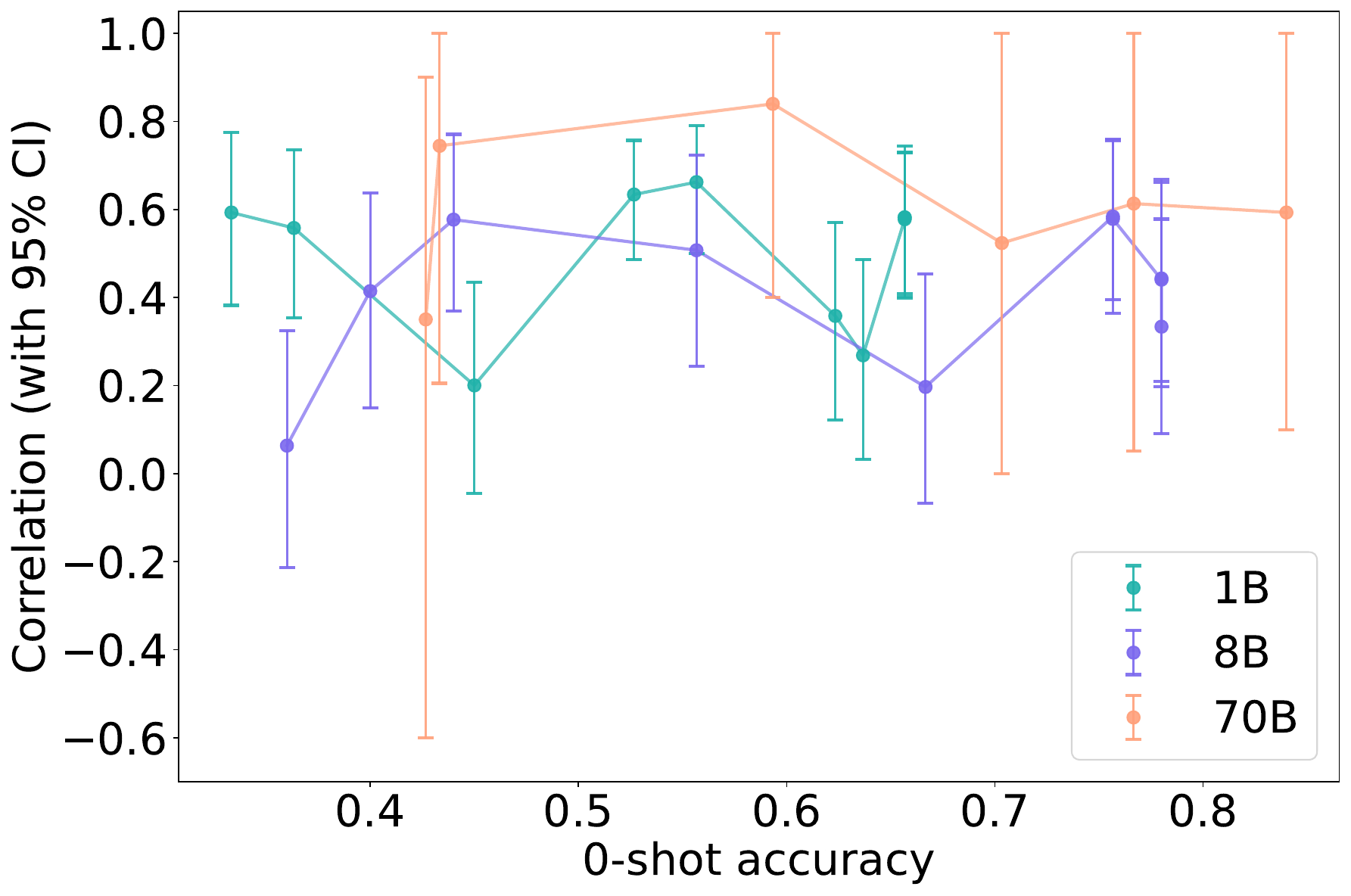}
        \caption{3-way classification}
        \label{fig:horizontal_correlations_3classes}
    \end{subfigure}
    \begin{subfigure}{0.4\textwidth}
        \centering
        \includegraphics[width=\linewidth]{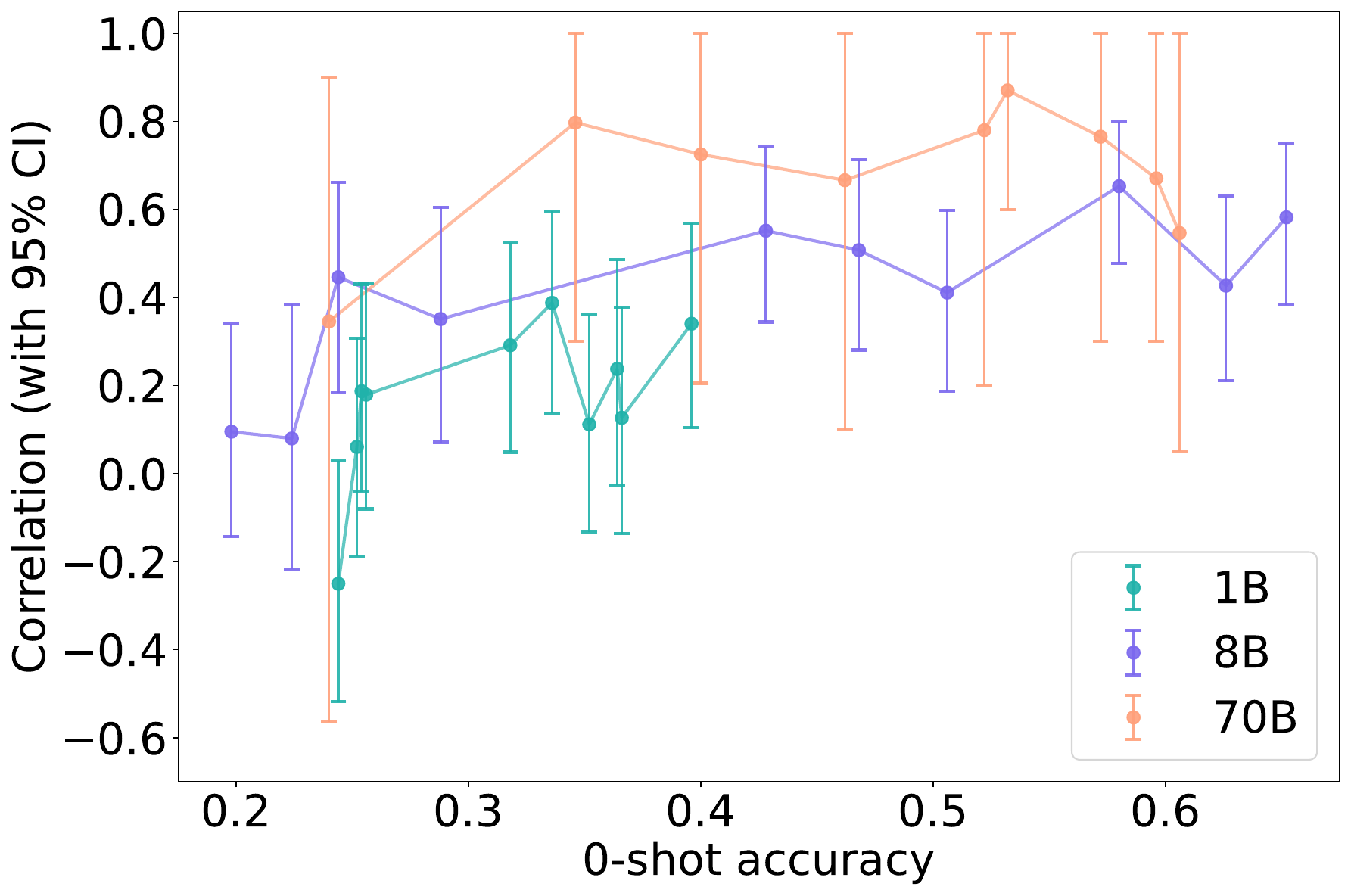}
        \caption{5-way classification}
        \label{fig:horizontal_correlations_5classes}
    \end{subfigure}

    \caption{\textbf{Evaluation of learning curves for label sets obtained with different $K$ labeling examples.} Ranking correlation coefficient between $N$ and $N$-shot accuracy vs. zero-shot accuracy for each curve. $N \in \{\text{num classes},...40\}$ for 1B and 8B models, $N \in \{10, 20, 30, 40\}$ for 70B model. Higher correlation indicates that the accuracy for that curve is often strictly increasing with $N$ (steeper curve), while lower accuracy indicates that the accuracy can be plateauing or decreasing on some intervals (flatter curve). The CI are computed over 1000 bootstrapping samples from 10
runs per N-shot accuracy.}
    \label{fig:horizontal_correlations}
\end{figure}

\paragraph{Model size influences the learning rate.} 

From Figure~\ref{fig:main_result} we observe that most learning curves are increasing. In Figure~\ref{fig:horizontal_correlations} we show that the slope depends on model size and zero-shot accuracy. The larger 70B model is more efficient; it makes more use of fewer examples and thus exhibits steeper curves (such as Figure~\ref{fig:accuracy_3classes}, 70B model, $K=30$). The $N$-shot accuracies for this curve are highly correlated with $N$ (see Figure~\ref{fig:horizontal_correlations_3classes} orange curve, zero-shot accuracy 59\%). With representations of a similar zero-shot accuracy (40\%-60\% range), the smaller models can also learn, but their curves increase more slowly (and thus have a lower correlation between $N$ and $N$-shot accuracy), suggesting that it would take many more demonstrations to attain the same accuracy that the 70B model achieves with 20-30 demonstrations. 

\paragraph{Learning is conditioned by representation.}

Most of the learning curves typically increase, but there is a lot of variance in how much ICL improves with more demonstrations. The increase between the minimum accuracy (zero-shot) and the maximum accuracy (40-shot) ranges from $0\%$ up to $25\%$. We observe that the representations fall into three categories: small, medium, and high zero-shot accuracy. 
The small, zero-shot accuracy representations are usually found with a small $K$ number of labeling examples and are not intuitive or appropriate names for the task. This type of label set makes the task challenging: the model may have to infer the true nature of the task (possibly by inferring more suitable class names) and then map the unintuitive labels onto them. It is not always apparent from the sentences that they illustrate a sentiment classification task. For example, a sentence like ``\textit{In the upcoming season, I'll be in the zone every time I step onto the court.}'' labeled with ``\textit{cloudy},'' might distract the model from the clustering of sentences into appropriate classes. Typically for representations like this, the models start with near-chance zero-shot accuracy, and the accuracy increases only very little regardless of how many demonstrations are presented (e.g. Figure~\ref{fig:accuracy_3classes} 8B model, $K=10$ and Figure~\ref{fig:accuracy_5classes} 8B model, $K \in \{10, 20\}$). The representations with a medium (40\%--60\%) zero-shot accuracy benefit the most from demonstrations. They can get 15\%--25\% improvement from the baseline by seeing demonstrations. These labels are sufficiently suggestive of the task \{\textit{medically, offending, celebrations}\} that the model can eventually determine the mapping. 

The last group of representations consists of the high zero-shot accuracy representations, those that match or are very close to gold labels. These label sets are already close to the ceiling accuracy possible for each model size. For instance, in Figure~\ref{fig:accuracy_3classes} the 1B model with $K\in\{60, 70\}$ starts at 58\%, and plateaus at 62\%, the 8B model with $K\in\{80, 90, 100\}$ starts at 77\% and goes to 82\%, and the 70B model with $K\in\{80, 90, 100\}$ starts at 84\% and plateaus at 87\%. In this group, we observed one exception. In Figure~\ref{fig:accuracy_3classes}, for 3-way classification with a 1B model, the curve corresponding to $K\in{80, 90, 100}$ initially decreases before increasing. One of the labels in this set is the translation of the word \textit{danger} in Nepali. The ICL task may be harder because it requires multilingual reasoning, which can involve translation as a first step before figuring out the input-output mapping. It appears that for $N < 9$ examples, the model is confused and thus the accuracy decreases, but it quickly recovers and achieves a high accuracy toward $N=40$ demonstrations, as expected for the corresponding zero-shot accuracy.

\section{Conclusion}

The success of ICL has previously been attributed to how the in-context demonstrations are represented, and prior work has questioned whether true learning is, in fact, happening~\citep{NEURIPS2021_5c049256, min-etal-2022-rethinking}. Previous observations show that ICL performance improves with the number of demonstrations for both gold and abstract labels~\citep{kirsanov-etal-2025-geometry}, with gold labels consistently outperforming abstract ones. Based on this, we hypothesized that the choice of representation influences the learning trajectory in ICL. We developed an algorithm to enumerate a spectrum of label representations varying in semantic relevance and tested the performance of these label sets in ICL. We found that the representation of demonstrations determines the baseline accuracy of ICL, as measured by zero-shot performance. The relative quality of the label sets is consistent across demonstrations, and follows the order determined by the baseline accuracies. Furthermore, this baseline typically limits the range of attainable accuracies. It is possible to overcome the limits of the representation, but only with a large number of demonstrations and larger models. 
The efficiency of learning, measured as the slope of improvement over in-context demonstrations, is influenced both by the quality of representation and model size. Representations with a medium zero-shot accuracy typically benefit the most from seeing more demonstrations and have a higher slope, and larger models can learn faster. 
In summary, our work reveals the relationship between number of demonstrations and how they are represented on ICL performance, and highlights the importance of considering the representation when studying properties of in-context learning from demonstrations.

Our findings on the interaction between learning and representation in LLMs 
closely reflect what we know of more conventional neural network learning.
% are a natural extension of traditional machine learning knowledge. 
The search for high-performing prompts for LLMs is in spirit similar to hyperparameter search \citep{Bengio+chapter2007, liu2018darts} for neural network classifiers that learn via backpropagation. \cite{NEURIPS2021_5c049256} found that good prompts are effective because they are chosen using large validation sets. The prompts influence the model behavior similarly to how a choice of initialization influences neural network training. In particular, the choice of label representation in ICL is analogous to the feature selection for the inputs of a neural network classifier. The different choices of representation determine the learning trajectory in both cases: for LLMs, a high quality representation leads to a high zero-shot accuracy and faster convergence; for neural network classifiers, a good set of features can lead to efficient learning~\citep{LeCun2012}.

Beyond in-context learning, LLMs have shown high performance on complex reasoning tasks, such as programming and mathematical problem solving \citep{Guo2025, ruis2025procedural}. Our study also has potential implications about the role of representation in such reasoning. The finding that the representation determines both a baseline accuracy and the efficiency of in-context learning suggests that LLMs already have useful priors, but in order to make the most use of them, we need to present the task in an appropriate manner. Extending these findings about ICL to more complex reasoning tasks could offer a more nuanced understanding about memorization vs. reasoning in LLMs \citep{bowen-etal-2024-comprehensive, jin-etal-2025-disentangling-memory, salido2025othersgeneraltechniquedistinguish}. Moreover, our findings could explain LLM reasoning failures when changing parameters of an original problem such as document length or the number of variables in a math problem \citep{malek2025frontierllmsstrugglesimple}. Such changes in the prompt, despite attempting to preserve the fundamental difficulty of a problem, result in a significant change in the representation, which lowers the baseline accuracy.

\subsubsection*{Acknowledgments}

This work was partly supported by the Institute of Information \& Communications Technology Planning \& Evaluation (IITP) with a grant funded by the Ministry of Science and ICT (MSIT) of the Republic of Korea in connection with the Global AI Frontier Lab International Collaborative Research,  the Samsung Advanced Institute of Technology (under the project Next Generation Deep Learning: From Pattern Recognition to AI) and the National Science Foundation (under NSF Award 1922658).

\bibliography{iclr2026_conference}
\bibliographystyle{iclr2026_conference}

\newpage
\clearpage
\appendix
\section{List of label sets}
\label{appendix:label_sets}

\begin{table}[h!]
\centering
\begin{tabular}{c|c|c|c}
\hline
\textbf{K} & \textbf{1B} & \textbf{8B} & \textbf{70B} \\
\hline
10  & \makecell{Nutrition, Giz \\ Legends} & \makecell{Gluten, Laptop \\ clouds} & \makecell{biomedical, malware \\ cloudy} \\ \hline
20  & \makecell{diabetes, Hacker \\ Presbyterian} & \makecell{Diabetes, Revenge \\ spirit} & \makecell{fitness, computer \\ joyful} \\ \hline
30  & \makecell{overweight, annoy \\ scholarships} & \makecell{FDA, console \\ celebration} & \makecell{Obesity, rage \\ celebration} \\ \hline
40  & \makecell{medically, offending \\ celebrating} & \makecell{fearful, malicious \\ celebration} & \makecell{panic, rage \\ Cheers} \\ \hline
50  & \makecell{medically, offending \\ celebrations} & \makecell{digestive, insulting \\ accomplishments} & \makecell{panic, rage \\ Cheers} \\ \hline
60  & \makecell{panicked, offending \\ celebrations} & \makecell{fears, insults \\ joyful} & \makecell{worry, complain \\ celebration} \\ \hline
70  & \makecell{hazardous, offending \\ celebrations} & \makecell{fears, insults \\ joyful} & \makecell{fear, angry \\ happy} \\ \hline
80  & \makecell{à¤kà¤¤à¤° (\textit{danger}, Nepali), offending \\ celebrations} & \makecell{fears, complaints \\ joyful} & \makecell{fear, angry \\ happy} \\ \hline
90  & \makecell{à¤kà¤¤à¤° (\textit{danger}, Nepali), offending \\ celebrations} & \makecell{fears, complaints \\ joyful} & \makecell{fear, angry \\ happy} \\ \hline
100 & \makecell{à¤kà¤¤à¤° (\textit{danger}, Nepali), offending \\ celebrations} & \makecell{fears, complaint \\ joyful} & \makecell{fear, angry \\ happy} \\ \hline
\end{tabular}
\caption{Label sets obtained from running Algorithm~\ref{alg:hill_climb} on $K$ labeling examples for 3-way classification}
\label{tab:labels_3classes}
\end{table}

\begin{table}[h!]
\centering
\begin{tabular}{c|c|c|c}
\hline
\textbf{K} & \textbf{1B} & \textbf{8B} & \textbf{70B} \\
\hline
10  & \makecell{movie, Musik \\ Causes, Roller \\ NRL} & \makecell{theater, COLOR \\ HEALTH, ride \\ Offensive} & \makecell{Marvel, MUSIC \\ HEALTH, roller \\ veh} \\ \hline
20  & \makecell{witches, audition \\ bere, adip \\ Messi} & \makecell{cinema, Broadcasting \\ Deng, nut \\ Rugby} & \makecell{Magical, positive \\ loss, Dietary \\ Baseball} \\ \hline
30  & \makecell{trick, Dresses \\ bere, hysteria \\ Messi} & \makecell{surprising \\ MÃ©d (\textit{Med}, French) \\ ÑģÐ¾Ð½ (\textit{dream}, Russian) \\ snack, soccer} & \makecell{surprise, celebration \\ tragedy, amused \\ ìĬ¤íı¬ì¸ł (\textit{sports}, Korean)} \\ \hline
40  & \makecell{puzzle, Ventures \\ à¤h (\textit{A}, Marathi), carniv \\ Penalty} & \makecell{surprising \\ ÑĤÐµÐ» (\textit{tel}, Russian) \\ resignation, aliment \\ soccer} & \makecell{surprising, positives \\ heartbreaking \\ Ð¿Ð¸Ñī (\textit{shout}, Bulgarian)\\ frustrated} \\ \hline
50  & \makecell{spectacle, talent \\ mourn, endanger \\ offense} & \makecell{surprised, baÅŁarÄ± \\(\textit{success}, Turkish)\\ sadness, xen, Rage} & \makecell{surprising, positives \\ heartbreaking, Brussels \\ frustrated} \\ \hline
60  & \makecell{spectacle, production \\ mourn, peril \\ offense} & \makecell{amazed, Ð½Ð°ÑĥÐº \\ (\textit{science}, Ukrainian) \\ mourn, scare, brawl} & \makecell{surprising, positives \\ heartbreaking, nerv \\ agg} \\ \hline
70  & \makecell{spectacle, productions \\ mourn, peril \\ racket} & \makecell{surprising, íh (\textit{?})\\ sorrow, terror \\ hostile} & \makecell{surprise, pleasant \\ sorrow, fears \\ rage} \\ \hline
80  & \makecell{spectacle, dance \\ mourn, peril \\ criticizing} & \makecell{surprising, celebrates \\ condolences, terror \\ rage} & \makecell{surprise, pleasant \\ sorrow, fears \\ rage} \\ \hline
90  & \makecell{magician, dancer \\ mourning, risking \\ wrath} & \makecell{surprising, joyful \\ sorrow, fears \\ rage} & \makecell{surprise, Lift \\ broken, fears \\ rage} \\ \hline
100 & \makecell{spectacle, dance \\ condolences, peril \\ pissed} & \makecell{surprising, joyful \\ sorrow, fears \\ anger} & \makecell{surprise, happy \\ sad, anxious \\ ang} \\ \hline
\end{tabular}
\caption{Label sets obtained from from running Algorithm~\ref{alg:hill_climb} on $K$ labeling examples for 5-way classification}
\label{tab:labels_5classes}
\end{table}

\newpage
\clearpage

\section{Learning curves}
\label{appendix:100demos}
We show the full raw (unsmoothed) learning curves for up to 100 demonstrations for 1B and 8B models for 3-way classification in Figure~\ref{fig:3-way_raw_curves} and for 5-way classification in Figure~\ref{fig:5-way_raw_curves}.

\begin{figure}[ht]
    \centering
    
    \begin{subfigure}{\linewidth}
        \centering
        \includegraphics[width=\linewidth]{figures/plots_single/accuracy_curves_1b_3classes.pdf}
        %\caption{1B 3 classes}
        \label{fig:1B_3classes}
    \end{subfigure}
    
    \vspace{1em} % space between subfigures
    
    \begin{subfigure}{\linewidth}
        \centering
        \includegraphics[width=\linewidth]{figures/plots_single/accuracy_curves_8b_3classes.pdf}
        %\caption{8B 3 classes}
        \label{fig:8B_3classes}
    \end{subfigure}
    
    \caption{3-way classification}
    \label{fig:3-way_raw_curves}
\end{figure}

\begin{figure}[ht]
    \centering
    
    \begin{subfigure}{\linewidth}
        \centering
        \includegraphics[width=\linewidth]{figures/plots_single/accuracy_curves_1b_5classes.pdf}
        %\caption{1B 3 classes}
        \label{fig:1B_5classes}
    \end{subfigure}
    
    \vspace{1em} % space between subfigures
    
    \begin{subfigure}{\linewidth}
        \centering
        \includegraphics[width=\linewidth]{figures/plots_single/accuracy_curves_8b_5classes.pdf}
        %\caption{8B 3 classes}
        \label{fig:8B_5classes}
    \end{subfigure}
    
    \caption{5-way classification}
    \label{fig:5-way_raw_curves}
\end{figure}

\newpage
\clearpage

\section{Correlation statistics from Figure ~\ref{fig:vertical_correlations}}

\begin{table}[ht]
\centering
\begin{tabular}{rccccc}
\hline
$n_{\text{demo}}$ & Mean Corr. & Std Corr. & Median Corr. & CI 2.5\% & CI 97.5\% \\
\hline
3  & 0.5486 & 0.1828 & 0.5723 & 0.1160 & 0.8336 \\
4  & 0.6087 & 0.1933 & 0.6308 & 0.1534 & 0.9006 \\
5  & 0.4634 & 0.2265 & 0.4970 & -0.0436 & 0.8320 \\
6  & 0.3776 & 0.2408 & 0.4056 & -0.1363 & 0.8012 \\
7  & 0.5743 & 0.1666 & 0.5908 & 0.2154 & 0.8582 \\
8  & 0.4798 & 0.2303 & 0.5000 & 0.0178 & 0.8493 \\
9  & 0.4261 & 0.1882 & 0.4356 & 0.0307 & 0.7655 \\
10 & 0.4111 & 0.2364 & 0.4246 & -0.1043 & 0.8185 \\
11 & 0.3826 & 0.2127 & 0.4062 & -0.0926 & 0.7447 \\
12 & 0.5043 & 0.2660 & 0.5280 & -0.0620 & 0.8863 \\
13 & 0.6692 & 0.1574 & 0.6770 & 0.3323 & 0.9293 \\
14 & 0.5197 & 0.1977 & 0.5354 & 0.1040 & 0.8530 \\
15 & 0.5170 & 0.1762 & 0.5215 & 0.1420 & 0.8431 \\
16 & 0.4968 & 0.1781 & 0.4985 & 0.1169 & 0.8185 \\
17 & 0.4846 & 0.1518 & 0.4862 & 0.1531 & 0.7693 \\
18 & 0.5639 & 0.1631 & 0.5706 & 0.2400 & 0.8739 \\
19 & 0.4597 & 0.1964 & 0.4653 & 0.0431 & 0.8037 \\
20 & 0.6816 & 0.1389 & 0.6893 & 0.3950 & 0.9171 \\
21 & 0.5907 & 0.1245 & 0.5846 & 0.3620 & 0.8510 \\
22 & 0.6074 & 0.1372 & 0.6074 & 0.3508 & 0.8800 \\
23 & 0.5401 & 0.1593 & 0.5461 & 0.2025 & 0.8406 \\
24 & 0.5110 & 0.1682 & 0.5215 & 0.1657 & 0.8089 \\
25 & 0.5436 & 0.1768 & 0.5583 & 0.1533 & 0.8529 \\
26 & 0.6441 & 0.1377 & 0.6442 & 0.3754 & 0.8896 \\
27 & 0.5369 & 0.2018 & 0.5461 & 0.1043 & 0.8897 \\
28 & 0.5540 & 0.1569 & 0.5539 & 0.2277 & 0.8677 \\
29 & 0.5478 & 0.1135 & 0.5354 & 0.3642 & 0.7939 \\
30 & 0.5293 & 0.1186 & 0.5338 & 0.3374 & 0.7694 \\
31 & 0.6273 & 0.1472 & 0.6319 & 0.3252 & 0.8896 \\
32 & 0.5963 & 0.1561 & 0.6031 & 0.2345 & 0.8923 \\
33 & 0.6415 & 0.1312 & 0.6442 & 0.3865 & 0.8800 \\
34 & 0.6192 & 0.1265 & 0.6197 & 0.4000 & 0.8677 \\
35 & 0.7148 & 0.1223 & 0.7200 & 0.4492 & 0.9142 \\
36 & 0.7034 & 0.1209 & 0.7178 & 0.4479 & 0.9047 \\
37 & 0.6910 & 0.1282 & 0.7017 & 0.4320 & 0.9047 \\
38 & 0.5380 & 0.1757 & 0.5556 & 0.0983 & 0.8308 \\
39 & 0.5583 & 0.1215 & 0.5516 & 0.3508 & 0.8062 \\
40 & 0.6259 & 0.1396 & 0.6339 & 0.3395 & 0.8678 \\
\hline
\end{tabular}
\caption{3-way classification, 1B model (green curve in Figure ~\ref{fig:vertical_correlations_3classes}). Ranking correlations across label sets for different numbers of demonstrations $n_{\text{demo}}$ (bootstrap = 1000 samples).}
\label{tab:vertical_correlations_1B_3classes}
\end{table}

\begin{table}[ht]
\centering
\begin{tabular}{rccccc}
\hline
$n_{\text{demo}}$ & Mean Corr. & Std Corr. & Median Corr. & CI 2.5\% & CI 97.5\% \\
\hline
3  & 0.8484 & 0.0735 & 0.8528 & 0.6892 & 0.9663 \\
4  & 0.8480 & 0.0822 & 0.8568 & 0.6647 & 0.9725 \\
5  & 0.7627 & 0.1246 & 0.7778 & 0.4628 & 0.9540 \\
6  & 0.7723 & 0.1528 & 0.7898 & 0.4204 & 0.9816 \\
7  & 0.8324 & 0.0875 & 0.8396 & 0.6439 & 0.9724 \\
8  & 0.8713 & 0.0760 & 0.8841 & 0.6944 & 0.9847 \\
9  & 0.8937 & 0.0536 & 0.9013 & 0.7655 & 0.9754 \\
10 & 0.8945 & 0.0599 & 0.9013 & 0.7509 & 0.9785 \\
11 & 0.8859 & 0.0658 & 0.8924 & 0.7324 & 0.9847 \\
12 & 0.8623 & 0.0827 & 0.8797 & 0.6647 & 0.9754 \\
13 & 0.8540 & 0.0804 & 0.8616 & 0.6563 & 0.9754 \\
14 & 0.8846 & 0.0577 & 0.8890 & 0.7570 & 0.9754 \\
15 & 0.9063 & 0.0548 & 0.9136 & 0.7809 & 0.9847 \\
16 & 0.8931 & 0.0566 & 0.8986 & 0.7654 & 0.9847 \\
17 & 0.9045 & 0.0522 & 0.9109 & 0.7878 & 0.9816 \\
18 & 0.9061 & 0.0605 & 0.9164 & 0.7509 & 0.9847 \\
19 & 0.8844 & 0.0647 & 0.8948 & 0.7385 & 0.9847 \\
20 & 0.9155 & 0.0514 & 0.9232 & 0.7902 & 0.9847 \\
21 & 0.9117 & 0.0469 & 0.9170 & 0.8062 & 0.9816 \\
22 & 0.8878 & 0.0644 & 0.8924 & 0.7447 & 0.9847 \\
23 & 0.8936 & 0.0602 & 0.9013 & 0.7509 & 0.9754 \\
24 & 0.9147 & 0.0472 & 0.9229 & 0.8037 & 0.9847 \\
25 & 0.9026 & 0.0546 & 0.9109 & 0.7895 & 0.9816 \\
26 & 0.9153 & 0.0455 & 0.9226 & 0.8117 & 0.9847 \\
27 & 0.9582 & 0.0245 & 0.9630 & 0.8986 & 0.9877 \\
28 & 0.9381 & 0.0381 & 0.9478 & 0.8493 & 0.9877 \\
29 & 0.9282 & 0.0482 & 0.9398 & 0.8124 & 0.9877 \\
30 & 0.9086 & 0.0521 & 0.9136 & 0.8001 & 0.9877 \\
31 & 0.8965 & 0.0547 & 0.9011 & 0.7778 & 0.9847 \\
32 & 0.9002 & 0.0527 & 0.9072 & 0.7901 & 0.9847 \\
33 & 0.9498 & 0.0286 & 0.9507 & 0.8863 & 0.9877 \\
34 & 0.9187 & 0.0491 & 0.9232 & 0.8068 & 0.9847 \\
35 & 0.9206 & 0.0471 & 0.9260 & 0.8185 & 0.9847 \\
36 & 0.9046 & 0.0476 & 0.9109 & 0.8025 & 0.9754 \\
37 & 0.8970 & 0.0520 & 0.9013 & 0.7809 & 0.9754 \\
38 & 0.9239 & 0.0445 & 0.9291 & 0.8250 & 0.9877 \\
39 & 0.9192 & 0.0440 & 0.9259 & 0.8209 & 0.9847 \\
40 & 0.9086 & 0.0513 & 0.9136 & 0.7878 & 0.9847 \\
\hline
\end{tabular}
\caption{3-way classification, 8B model (purple curve in Figure ~\ref{fig:vertical_correlations_3classes}). Ranking correlations across label sets for different numbers of demonstrations $n_{\text{demo}}$ (bootstrap = 1000 samples).}
\label{tab:vertical_correlations_8B_3classes}
\end{table}

\begin{table}[ht]
\centering
\begin{tabular}{rccccc}
\hline
$n_{\text{demo}}$ & Mean Corr. & Std Corr. & Median Corr. & CI 2.5\% & CI 97.5\% \\
\hline
10 & 0.8732 & 0.0769 & 0.8857 & 0.7143 & 1.0000 \\
20 & 0.8978 & 0.0808 & 0.9429 & 0.7143 & 1.0000 \\
30 & 0.8771 & 0.1039 & 0.8986 & 0.5798 & 1.0000 \\
40 & 0.9108 & 0.0972 & 0.9429 & 0.6571 & 1.0000 \\
\hline
\end{tabular}
\caption{3-way classification, 70B model (orange curve in Figure ~\ref{fig:vertical_correlations_3classes}). Ranking correlations across label sets for different numbers of demonstrations $n_{\text{demo}}$ (bootstrap = 1000 samples).}
\label{tab:vertical_correlations_70B_3classes}
\end{table}

\begin{table}[ht]
\centering
\begin{tabular}{rccccc}
\hline
$n_{\text{demo}}$ & Mean Corr. & Std Corr. & Median Corr. & CI 2.5\% & CI 97.5\% \\
\hline
5  & 0.6375 & 0.1627 & 0.6444 & 0.2721 & 0.8910 \\
6  & 0.6621 & 0.2044 & 0.7016 & 0.1567 & 0.9387 \\
7  & 0.5193 & 0.2427 & 0.5636 & -0.0324 & 0.8571 \\
8  & 0.4579 & 0.2972 & 0.4817 & -0.1626 & 0.9030 \\
9  & 0.4426 & 0.2540 & 0.4788 & -0.1342 & 0.8390 \\
10 & 0.5284 & 0.2252 & 0.5710 & 0.0485 & 0.8573 \\
11 & 0.4324 & 0.2577 & 0.4602 & -0.1664 & 0.8303 \\
12 & 0.3863 & 0.2584 & 0.3988 & -0.1030 & 0.8477 \\
13 & 0.4129 & 0.2881 & 0.4479 & -0.2121 & 0.8788 \\
14 & 0.5727 & 0.2140 & 0.5957 & 0.0915 & 0.8998 \\
15 & 0.6749 & 0.1798 & 0.7091 & 0.2118 & 0.9180 \\
16 & 0.5898 & 0.2114 & 0.6140 & 0.1090 & 0.9152 \\
17 & 0.7062 & 0.1538 & 0.7333 & 0.3281 & 0.9362 \\
18 & 0.5889 & 0.2328 & 0.6371 & 0.0182 & 0.9119 \\
19 & 0.6046 & 0.1892 & 0.6322 & 0.1758 & 0.8875 \\
20 & 0.6725 & 0.1878 & 0.7052 & 0.2438 & 0.9268 \\
21 & 0.6258 & 0.1622 & 0.6575 & 0.2673 & 0.8754 \\
22 & 0.5416 & 0.2144 & 0.5593 & 0.0793 & 0.8875 \\
23 & 0.5537 & 0.2003 & 0.5888 & 0.0910 & 0.8633 \\
24 & 0.6124 & 0.1902 & 0.6242 & 0.1877 & 0.9030 \\
25 & 0.5332 & 0.2402 & 0.5394 & -0.0064 & 0.9030 \\
26 & 0.5958 & 0.2098 & 0.6353 & 0.1155 & 0.9067 \\
27 & 0.6985 & 0.1357 & 0.7091 & 0.3951 & 0.9119 \\
28 & 0.6916 & 0.1213 & 0.7052 & 0.4423 & 0.9030 \\
29 & 0.6130 & 0.1596 & 0.6252 & 0.2605 & 0.8754 \\
30 & 0.7262 & 0.1624 & 0.7660 & 0.3343 & 0.9329 \\
31 & 0.7641 & 0.1523 & 0.7939 & 0.3888 & 0.9606 \\
32 & 0.6992 & 0.1691 & 0.7333 & 0.3100 & 0.9394 \\
33 & 0.6159 & 0.1857 & 0.6444 & 0.1581 & 0.8997 \\
34 & 0.6727 & 0.1698 & 0.7052 & 0.2917 & 0.9119 \\
35 & 0.7016 & 0.1831 & 0.7576 & 0.3251 & 0.9483 \\
36 & 0.7466 & 0.1361 & 0.7697 & 0.4133 & 0.9391 \\
37 & 0.7843 & 0.1305 & 0.8146 & 0.4479 & 0.9545 \\
38 & 0.7434 & 0.1212 & 0.7538 & 0.4862 & 0.9484 \\
39 & 0.7288 & 0.1418 & 0.7516 & 0.4109 & 0.9394 \\
40 & 0.6682 & 0.1699 & 0.6930 & 0.2606 & 0.9119 \\
\hline
\end{tabular}
\caption{5-way classification, 1B model (green curve in Figure ~\ref{fig:vertical_correlations_5classes}). Ranking correlations across label sets for different numbers of demonstrations $n_{\text{demo}}$ (bootstrap = 1000 samples).}
\label{tab:vertical_correlations_1B_5classes}
\end{table}

\begin{table}[ht]
\centering
\begin{tabular}{rccccc}
\hline
$n_{\text{demo}}$ & Mean Corr. & Std Corr. & Median Corr. & CI 2.5\% & CI 97.5\% \\
\hline
5  & 0.8698 & 0.0758 & 0.8788 & 0.6969 & 0.9758 \\
6  & 0.9022 & 0.0546 & 0.9119 & 0.7669 & 0.9848 \\
7  & 0.9227 & 0.0437 & 0.9273 & 0.8061 & 0.9879 \\
8  & 0.9013 & 0.0593 & 0.9152 & 0.7576 & 0.9879 \\
9  & 0.8687 & 0.0740 & 0.8815 & 0.6969 & 0.9758 \\
10 & 0.9027 & 0.0550 & 0.9152 & 0.7573 & 0.9758 \\
11 & 0.8831 & 0.0754 & 0.9030 & 0.7054 & 0.9849 \\
12 & 0.8612 & 0.0773 & 0.8754 & 0.6809 & 0.9755 \\
13 & 0.9225 & 0.0406 & 0.9273 & 0.8303 & 0.9879 \\
14 & 0.9112 & 0.0581 & 0.9273 & 0.7697 & 0.9879 \\
15 & 0.9331 & 0.0353 & 0.9394 & 0.8545 & 0.9879 \\
16 & 0.9222 & 0.0419 & 0.9273 & 0.8207 & 0.9879 \\
17 & 0.9276 & 0.0353 & 0.9362 & 0.8509 & 0.9879 \\
18 & 0.9045 & 0.0469 & 0.9152 & 0.7939 & 0.9758 \\
19 & 0.8882 & 0.0604 & 0.9030 & 0.7333 & 0.9758 \\
20 & 0.9530 & 0.0275 & 0.9515 & 0.8908 & 0.9970 \\
21 & 0.9438 & 0.0370 & 0.9515 & 0.8510 & 0.9970 \\
22 & 0.9449 & 0.0306 & 0.9515 & 0.8788 & 0.9879 \\
23 & 0.9458 & 0.0308 & 0.9515 & 0.8788 & 0.9879 \\
24 & 0.9453 & 0.0263 & 0.9483 & 0.8875 & 0.9879 \\
25 & 0.9227 & 0.0354 & 0.9273 & 0.8449 & 0.9758 \\
26 & 0.9374 & 0.0345 & 0.9394 & 0.8667 & 0.9879 \\
27 & 0.9236 & 0.0393 & 0.9273 & 0.8303 & 0.9879 \\
28 & 0.9289 & 0.0378 & 0.9362 & 0.8424 & 0.9879 \\
29 & 0.9128 & 0.0468 & 0.9165 & 0.8060 & 0.9818 \\
30 & 0.9524 & 0.0249 & 0.9515 & 0.8909 & 0.9879 \\
31 & 0.9411 & 0.0311 & 0.9423 & 0.8667 & 0.9879 \\
32 & 0.9385 & 0.0332 & 0.9394 & 0.8667 & 0.9879 \\
33 & 0.9466 & 0.0257 & 0.9515 & 0.8909 & 0.9879 \\
34 & 0.9403 & 0.0297 & 0.9394 & 0.8788 & 0.9879 \\
35 & 0.9480 & 0.0295 & 0.9515 & 0.8788 & 0.9879 \\
36 & 0.9223 & 0.0392 & 0.9273 & 0.8424 & 0.9879 \\
37 & 0.9429 & 0.0281 & 0.9483 & 0.8788 & 0.9879 \\
38 & 0.9320 & 0.0414 & 0.9394 & 0.8292 & 0.9879 \\
39 & 0.9443 & 0.0306 & 0.9483 & 0.8788 & 0.9879 \\
40 & 0.9371 & 0.0315 & 0.9394 & 0.8667 & 0.9940 \\
\hline
\end{tabular}
\caption{5-way classification, 8B model (purple curve in Figure ~\ref{fig:vertical_correlations_5classes}). Ranking correlations across label sets for different numbers of demonstrations $n_{\text{demo}}$ (bootstrap = 1000 samples).}
\label{tab:vertical_correlations_8B_5classes}
\end{table}

\begin{table}[ht]
\centering
\begin{tabular}{rccccc}
\hline
$n_{\text{demo}}$ & Mean Corr. & Std Corr. & Median Corr. & CI 2.5\% & CI 97.5\% \\
\hline
10 & 0.8701 & 0.0744 & 0.8833 & 0.7000 & 0.9500 \\
20 & 0.8955 & 0.0476 & 0.9000 & 0.7500 & 0.9500 \\
30 & 0.8549 & 0.0776 & 0.8833 & 0.7000 & 0.9542 \\
40 & 0.8683 & 0.0730 & 0.8833 & 0.6946 & 0.9667 \\
\hline
\end{tabular}
\caption{5-way classification, 8B model (orange curve in Figure ~\ref{fig:vertical_correlations_5classes}). Ranking correlations across label sets for different numbers of demonstrations $n_{\text{demo}}$ (bootstrap = 1000 samples).}
\label{tab:vertical_correlations_70B_5classes}
\end{table}

\newpage
\clearpage
\section{Correlation statistics from Figure ~\ref{fig:horizontal_correlations}}

\begin{table}[ht]
\centering
\begin{tabular}{rccccc}
\hline
$K$ & Mean Corr. & Std Corr. & Median Corr. & CI 2.5\% & CI 97.5\% \\
\hline
\multicolumn{6}{c}{\textbf{1B}} \\
10  & 0.5931 & 0.0986 & 0.6029 & 0.3822 & 0.7745 \\
20  & 0.5579 & 0.0975 & 0.5615 & 0.3536 & 0.7347 \\
30  & 0.2003 & 0.1232 & 0.2018 & -0.0451 & 0.4343 \\
40  & 0.6339 & 0.0713 & 0.6428 & 0.4859 & 0.7570 \\
50  & 0.6622 & 0.0765 & 0.6675 & 0.4997 & 0.7909 \\
60  & 0.3582 & 0.1165 & 0.3626 & 0.1221 & 0.5709 \\
70  & 0.2686 & 0.1195 & 0.2741 & 0.0325 & 0.4863 \\
80  & 0.5777 & 0.0847 & 0.5819 & 0.4086 & 0.7279 \\
90  & 0.5821 & 0.0864 & 0.5867 & 0.3983 & 0.7435 \\
100 & 0.5818 & 0.0824 & 0.5873 & 0.4023 & 0.7308 \\
\hline
\multicolumn{6}{c}{\textbf{8B}} \\
10  & 0.0636 & 0.1384 & 0.0677 & -0.2137 & 0.3244 \\
20  & 0.4146 & 0.1236 & 0.4192 & 0.1486 & 0.6373 \\
30  & 0.5769 & 0.1038 & 0.5840 & 0.3687 & 0.7708 \\
40  & 0.1968 & 0.1308 & 0.1924 & -0.0673 & 0.4537 \\
50  & 0.5073 & 0.1236 & 0.5169 & 0.2436 & 0.7229 \\
60  & 0.5840 & 0.0941 & 0.5892 & 0.3956 & 0.7597 \\
70  & 0.5785 & 0.1011 & 0.5875 & 0.3644 & 0.7570 \\
80  & 0.4405 & 0.1200 & 0.4424 & 0.1980 & 0.6622 \\
90  & 0.4433 & 0.1153 & 0.4448 & 0.2089 & 0.6690 \\
100 & 0.3338 & 0.1289 & 0.3373 & 0.0912 & 0.5783 \\
\hline
\multicolumn{6}{c}{\textbf{70B}} \\
10  & 0.7446 & 0.2272 & 0.8000 & 0.2052 & 1.0000 \\
20  & 0.3503 & 0.4338 & 0.3000 & -0.6000 & 0.9000 \\
30  & 0.8398 & 0.1380 & 0.9000 & 0.4000 & 1.0000 \\
40  & 0.6134 & 0.2642 & 0.7000 & 0.0513 & 1.0000 \\
60  & 0.5237 & 0.2925 & 0.6000 & 0.0000 & 1.0000 \\
70  & 0.5931 & 0.2530 & 0.6156 & 0.1000 & 1.0000 \\
\hline
\end{tabular}
\caption{3-way classification correlations (between N and $N$-shot accuracy) from Figure~\ref{fig:horizontal_correlations_3classes} for 1B (green curve), 8B (purple curve), and 70B (orange curve) models across different $K$ values (bootstrap = 1000 samples). Each $K$ value corresponds to a learning curve, which is determined by its zero-shot accuracy in the figure.}
\label{tab:horizontal_correlations_all_models_3classes}
\end{table}

\begin{table}[ht]
\centering
\begin{tabular}{rccccc}
\hline
$K$ & Mean Corr. & Std Corr. & Median Corr. & CI 2.5\% & CI 97.5\% \\
\hline
\multicolumn{6}{c}{\textbf{1B}} \\
10  & 0.1794 & 0.1300 & 0.1774 & -0.0803 & 0.4313 \\
20  & -0.2500 & 0.1429 & -0.2555 & -0.5172 & 0.0297 \\
30  & 0.1871 & 0.1247 & 0.1825 & -0.0409 & 0.4303 \\
40  & 0.0607 & 0.1257 & 0.0557 & -0.1881 & 0.3077 \\
50  & 0.1119 & 0.1278 & 0.1156 & -0.1320 & 0.3608 \\
60  & 0.3877 & 0.1190 & 0.3982 & 0.1376 & 0.5956 \\
70  & 0.2916 & 0.1202 & 0.2941 & 0.0488 & 0.5246 \\
80  & 0.3405 & 0.1178 & 0.3437 & 0.1047 & 0.5689 \\
90  & 0.1268 & 0.1308 & 0.1282 & -0.1364 & 0.3775 \\
100 & 0.2378 & 0.1352 & 0.2406 & -0.0256 & 0.4862 \\
\hline
\multicolumn{6}{c}{\textbf{8B}} \\
10  & 0.0798 & 0.1523 & 0.0830 & -0.2172 & 0.3850 \\
20  & 0.0953 & 0.1268 & 0.0940 & -0.1428 & 0.3399 \\
30  & 0.4461 & 0.1226 & 0.4552 & 0.1838 & 0.6608 \\
40  & 0.3511 & 0.1377 & 0.3550 & 0.0710 & 0.6043 \\
50  & 0.5517 & 0.1062 & 0.5585 & 0.3444 & 0.7421 \\
60  & 0.5076 & 0.1090 & 0.5127 & 0.2809 & 0.7131 \\
70  & 0.4112 & 0.1039 & 0.4144 & 0.1865 & 0.5979 \\
80  & 0.6527 & 0.0810 & 0.6579 & 0.4772 & 0.7990 \\
90  & 0.5821 & 0.0931 & 0.5829 & 0.3833 & 0.7503 \\
100 & 0.4269 & 0.1062 & 0.4301 & 0.2116 & 0.6296 \\
\hline
\multicolumn{6}{c}{\textbf{70B}} \\
10  & 0.3457 & 0.3613 & 0.4000 & -0.5643 & 0.9000 \\
20  & 0.7973 & 0.1862 & 0.9000 & 0.3000 & 1.0000 \\
30  & 0.7250 & 0.1941 & 0.8000 & 0.2051 & 1.0000 \\
40  & 0.7799 & 0.1960 & 0.8000 & 0.2000 & 1.0000 \\
50  & 0.8705 & 0.1301 & 0.9000 & 0.6000 & 1.0000 \\
60  & 0.7653 & 0.1697 & 0.7000 & 0.3000 & 1.0000 \\
70  & 0.6708 & 0.2027 & 0.7000 & 0.3000 & 1.0000 \\
90  & 0.6664 & 0.2613 & 0.7000 & 0.1000 & 1.0000 \\
100 & 0.5465 & 0.2877 & 0.6000 & 0.0513 & 1.0000 \\
\hline
\end{tabular}
\caption{5-way classification correlations from Figure~\ref{fig:horizontal_correlations_5classes} for 1B (green curve), 8B (purple curve), and 70B (orange curve) models across different $K$ values (bootstrap = 1000 samples). Each $K$ value corresponds to a learning curve, which is determined by its zero-shot accuracy in the figure.}
\label{tab:horizontal_correlations_all_models_5classes}
\end{table}

\end{document}